\pdfoutput=1
\documentclass[a4paper,twoside]{article}

\usepackage{graphicx}
\usepackage{epsfig}
\usepackage{subcaption}
\usepackage{calc}
\usepackage{amssymb}
\usepackage{amstext}
\usepackage{amsmath}
\usepackage{amsthm}
\usepackage{multicol}
\usepackage{pslatex}
\usepackage{apalike}
\usepackage{algorithm2e}
\usepackage[bottom]{footmisc}
\usepackage{SCITEPRESS}     % SCITEPRESS format package
\usepackage{booktabs}
\usepackage{color}
\usepackage{xcolor}

\renewcommand{\paragraph}[1]{\noindent\textbf{#1}\hspace{2ex}}

\begin{document}

\title{End-to-End Optimization of \\Polarimetric Measurement and Material Classifier}

\author{\authorname{Ryota Maeda*\sup{1}\orcidAuthor{0000-0002-2407-5520}, 
Naoki Arikawa*\sup{1}, 
Yutaka No\sup{1} and 
Shinsaku Hiura\sup{1}\orcidAuthor{0000-0003-3176-097X}}
\affiliation{\sup{1}Graduate School of Engineering, University of Hyogo, Himeji, Japan}
\email{maeda.ryota.elerac@gmail.com, hiura@eng.u-hyogo.ac.jp}
}


\abstract{Material classification is a fundamental problem in computer vision and plays a crucial role in scene understanding. Previous studies have explored various material recognition methods based on reflection properties such as color, texture, specularity, and scattering. Among these cues, polarization is particularly valuable because it provides rich material information and enables recognition even at distances where capturing high-resolution texture is impractical.
However, measuring polarimetric reflectance properties typically requires multiple modulations of the polarization state of the incident light, making the process time-consuming and often unnecessary for certain recognition tasks. While material classification can be achieved using only a subset of polarimetric measurements, the optimal configuration of measurement angles remains unclear.
In this study, we propose an end-to-end optimization framework that jointly learns a material classifier and determines the optimal combinations of rotation angles for polarization elements that control both the incident and reflected light states. Using our Mueller-matrix material dataset, we demonstrate that our method achieves high-accuracy material classification even with a limited number of measurements.}

\onecolumn 
\maketitle 
\def\thefootnote{*}\footnotetext{Equal contribution}\def\thefootnote{\arabic{footnote}}
\normalsize 
\setcounter{footnote}{0}
\vfill

\section{Introduction}

Material recognition using cameras is a crucial task of visual understanding for automation in everyday environments. Conventional approaches that rely on color and texture cues from standard RGB cameras are widely used~\cite{caputo2005class}. However, their capability is inherently limited because these methods depend solely on RGB information and require capturing high-resolution texture. 

To overcome these limitations, previous works have explored a variety of optical cues beyond color and texture, including multispectral imaging~\cite{sato2015spectrans}\cite{saragadam2020programmable}, near-infrared~\cite{salamati2009material}, thermal imaging~\cite{saponaro2015material}\cite{dashpute2023thermal}, bidirectional reflectance distribution function (BRDF) measurements under varying illumination directions~\cite{wang2009material}\cite{jehle2010learning}\cite{liu2013discriminative}, and subsurface scattering~\cite{su2016material}\cite{tanaka2018material}\cite{lee2025spectral}\cite{pratama2025material}.
Among these modalities, polarization has proven to be particularly valuable for material recognition~\cite{hyde2010material}, and this study focuses on material recognition based on polarimetric reflectance properties.

The relationship between the incident and reflected polarization states is described by the Mueller matrix~\cite{collett2005field}, which consists of 16 parameters. Although it contains rich polarimetric information, measuring the full matrix is time-consuming because it requires rotating polarization elements on both the illumination and observation sides and capturing multiple images for various combinations of rotation angles. When using linear polarizers (LPs) and quarter-wave plates (QWPs), more than 20 combinations are typically required~\cite{baek2020image}, making the measurement cost a critical bottleneck for practical polarimetric material recognition.

However, complete Mueller matrix information is often unnecessary for material recognition. The polarimetric behavior of many natural materials can be represented by a small number of basis. For example, Fresnel reflection can be described using only three parameters: perpendicular (s-polarized) reflectance, parallel (p-polarized) reflectance, and phase delay~\cite{collett2005field}. This suggests that high recognition accuracy may be achievable with fewer, optimally chosen measurements, if informative polarization states can be identified. Yet, the optimal combination of polarization element rotation angles for this purpose remains unclear.

In this study, we propose a material classification method based on end-to-end joint optimization of the rotation angles in the polarimetric measurement system and the material classifier. This framework enables high-accuracy recognition with a reduced number of measurements.
Our work targets integration with LiDAR systems, allowing \textbf{independent classification at each measurement point without relying on texture information}.
To achive this approach, we construct a Mueller-matrix dataset of polarimetric reflectance properties for a diverse set of materials. Experimental results show that high classification accuracy can be achieved with only a few measurements across three measurement configurations with varying modulation complexity.

The main contributions of this work are summarized as follows:
\begin{itemize}
\item We propose a material classification method based on polarimetric reflectance properties that operates independently at each measurement point.
\item We achieve high-accuracy material classification with a reduced number of measurements through the simultaneous optimization of polarization element rotation angles and a material classifier.
\item We construct a comprehensive dataset of polarimetric reflectance properties covering a wide range of materials.
\end{itemize}

%%%
%偏光反射特性を詳細に取得するには，対象物体への入射光の偏光状態を変調することが有効である．そこで，LiDAR等と同様に，対象物体への入射光は偏光計測系により任意に変調可能であることを前提とする．

%直線偏光のみを用いる場合と，直線偏光と円偏光を利用する場合の両条件においてそれぞれ実験を行い，
% 本研究の貢献を以下に示す．
% \begin{itemize}
%     \item 偏光反射特性を用いた材質認識の提案
%     \item 偏光反射特性データセットの構築とそのデータの詳細な分析．
%     \item 偏光反射特性の冗長性に着目した，偏光計測系と分類器の同時最適化による，効率的な計測方法．
% \end{itemize}

\section{Related Work}

\paragraph{Material Recognition Using Diverse Light Information.}
Various physical properties of light have been leveraged to characterize material-specific reflectance behavior for material recognition. Prior studies have explored different modalities, including multispectral imaging in the visible and near-infrared regions~\cite{sato2015spectrans}\cite{saragadam2020programmable}\cite{salamati2009material}, thermal imaging~\cite{saponaro2015material}\cite{dashpute2023thermal}, bidirectional reflectance distribution function (BRDF) analysis~\cite{wang2009material}\cite{jehle2010learning}\cite{liu2013discriminative}, and subsurface scattering~\cite{su2016material}\cite{tanaka2018material}\cite{lee2025spectral}\cite{pratama2025material}.
Although these methods effectively capture rich reflectance information, acquiring comprehensive measurements typically require high costs in terms of both equipment and acquisition time.

To maintain recognition accuracy while reducing measurement costs, recent studies have emphasized measuring only the most informative components for material recognition. Liu et al.~\cite{liu2013discriminative} performed material classification using BRDF and wavelength information by optimizing illumination patterns across multiple light directions. Similarly, Saragadam et al.~\cite{saragadam2020programmable} identified optimal wavelength filter combinations using programmable spectral filters. Following this trend toward measurement-efficient recognition, our study focuses on optimizing the rotation angles of polarization elements attached to both the light source and the camera for polarization-based material recognition.

\paragraph{Scene Analysis Using Polarimetric Reflectance Properties.}
The measurement setup used in this study, in which polarization elements are attached to both the light source and the camera, corresponds to ellipsometry, a well-established technique in optics for applications such as thin-film analysis~\cite{azzam1978photopolarimetric}. Recently, the rich polarimetric reflectance information obtainable via ellipsometry has been utilized in various computer vision and graphics applications, including 3D shape reconstruction~\cite{baek2018simultaneous}\cite{hwang2022sparse}\cite{scheuble2024polarization}, polarimetric BRDF measurement~\cite{baek2020image}\cite{baek2022all}, light transport analysis~\cite{baek2021polarimetric}, and photoelasticity analysis~\cite{dave2025nest}.
However, acquiring such detailed polarimetric data requires multiple captures under varying polarization configurations, leading to considerable measurement overhead.

To mitigate this limitation, Maeda et al.~\cite{maeda2025event} achieved high-speed polarimetric reflectance measurement by combining rapidly rotating polarization elements with an event camera, though this approach requires specialized hardware. Baek et al.~\cite{baek2021polarimetric}\cite{baek2022all} proposed optimizing polarization element angles for more efficient Mueller matrix acquisition.
While our approach also involves angle optimization of polarization elements, it differs fundamentally by adopting an end-to-end learning framework: the measured light intensities are directly input to the material classifier without explicit Mueller matrix estimation. This integration enables simultaneous optimization of both the polarimetric measurement configuration and the material classification network.

\section{Polarimetric Reflectance Measurement}

% \subsection{Stokes Vectors and Mueller Matrices}
\subsection{Stokes-Mueller Calculus}
The polarization state of light and its transformations can be mathematically represented using Stokes vectors and Mueller matrices~\cite{collett2005field}. A Stokes vector $\mathbf{s}=[s_0, s_1, s_2, s_3]^\intercal \in \mathbb{R}^{4\times1}$ describes the polarization state of light, where $s_0$ represents the total light intensity, $s_1$ represents linear polarization at $0^{\circ}$ and $90^{\circ}$, $s_2$ represents linear polarization at $45^{\circ}$ and $135^{\circ}$, and $s_3$ represents circular polarization. These polarization states can be modified by polarizing optical elements or surface reflections. When $\mathbf{s}_\mathrm{in}$ represents the incident light and $\mathbf{s}_\mathrm{out}$ represents the reflected/transmitted light, their relationship can be expressed as $\mathbf{s}_\mathrm{out} = \mathbf{M} \mathbf{s}_\mathrm{in}$, where $\mathbf{M}\in\mathbb{R}^{4\times4}$ is the Mueller matrix. In this study, we use this Mueller matrix to represent polarimetric reflectance properties.

\subsection{Measurement of Polarimetric Reflectance Properties}
\label{sec:ellipsometer}
An ellipsometer is a method for measuring unknown polarimetric reflectance properties $\mathbf{M}$~\cite{collett2005field}. The ellipsometer illuminates the target object with light in multiple polarization states and estimates the Mueller matrix from the polarization states of the reflected light. Let $\mathbf{P}$ be the Mueller matrix of the polarization generator controlling the incident light's polarization state, $\mathbf{A}$ be the Mueller matrix of the polarization analyzer examining the reflected light's polarization state, and $\mathbf{s}=[L,0,0,0]$ be the Stokes vector of an unpolarized light source with intensity $L$. The detected light intensity $f$ can then be expressed as:
\begin{equation}
    f=\bigl[\,\mathbf{P} \mathbf{M} \mathbf{A} \mathbf{s}
    \,\bigr]_{0},
    \label{eq:ellipsometer}
\end{equation}
where $[...]_{0}$ denotes the extraction of the first component of the Stokes vector.

When performing $K$ measurements by varying the states of the polarization generator and analyzer, obtaining multiple polarization generator matrices $[\mathbf{P}_1, \mathbf{P}_2, \cdots, \mathbf{P}_K]$, analyzer matrices $[\mathbf{A}_1, \mathbf{A}_2, \cdots, \mathbf{A}_K]$, and light intensities $[f_1, f_2, \cdots, f_K]$, the unknown Mueller matrix $\mathbf{M}$ can be estimated by solving the following linear least squares problem:
\begin{equation}
    \underset{\mathbf{M}}{\text{minimize}} = \sum_{k=1}^{K} (f_k - [\mathbf{P}_k \mathbf{M} \mathbf{A}_k s]_{0})^2.
\end{equation}

\begin{figure}[t]
    \centering
    \includegraphics[width=0.99\hsize]{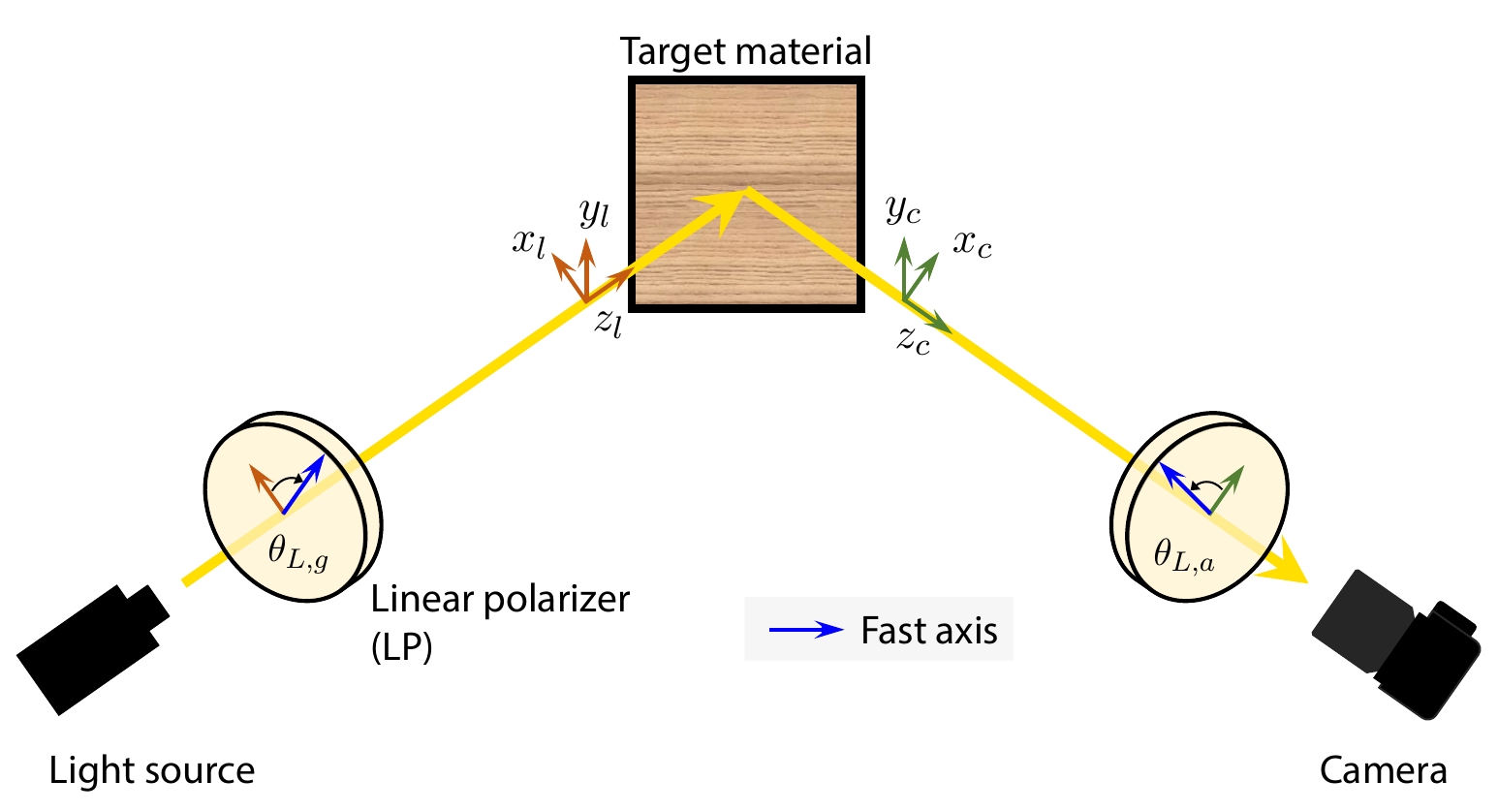}\\
    (a) Linear polarization measurement only
    \includegraphics[width=0.99\hsize]{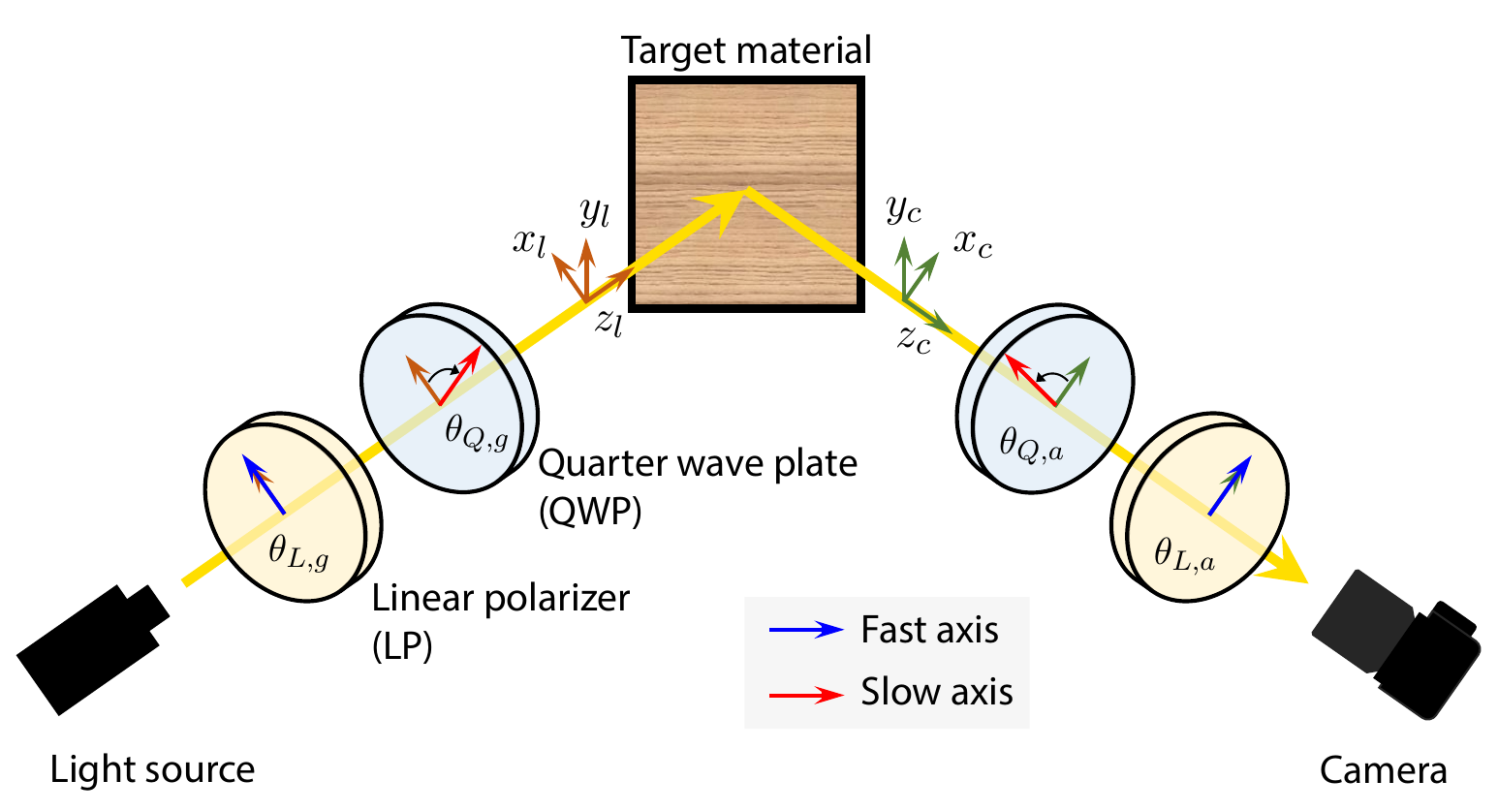} \\
    (b) Both linear and circular polarization measurements
    \caption{Optical setup for Mueller matrix measurement. Polarizing elements are positioned in front of both the light source and the camera. (a) For linear polarization components only, LPs are rotated. (b) For measuring all polarization states, including both linear and circular polarization, LPs are fixed while QWPs are rotated.}
%    \ecaption{Optical setup for measuring the Mueller matrix. Polarization optics are positioned in front of both the light source and the camera. (a) To measure only linear polarization, the linear polarizer is rotated. (b) To measure both linear and circular polarization states, the linear polarizers remain fixed while the quarter-wave plates are rotated.}
    \label{fig:polarization_measurement}
\end{figure}

While the polarization generator and analyzer can utilize any combination of polarizing elements, LPs and QWPs are commonly used~\cite{azzam1978photopolarimetric}\cite{baek2018simultaneous}\cite{baek2020image}. Figure~\ref {fig:polarization_measurement} shows the optical setup differences between measuring only linear polarization and measuring both linear and circular polarization.

\paragraph{Linear Polarization Only}
As shown in Figure~\ref{fig:polarization_measurement}(a), when measuring only linear polarization components, LPs are placed and rotated on both the light source and camera sides. When $\theta_{\mathrm{L},p}$ and $\theta_{\mathrm{L},a}$ represent the rotation angles of the LPs on the source and camera sides, respectively, $\mathbf{P}$ and $\mathbf{A}$ can be expressed as:
\begin{align}
    \mathbf{P} &= \mathbf{L}(\theta_{\mathrm{L},p}), \\
    \mathbf{A} &= \mathbf{L}(\theta_{\mathrm{L},a}),
\end{align}
where $\mathbf{L}$ is the Mueller matrix of a LP.

\paragraph{Both Linear and Circular Polarization}
As shown in Figure \ref{fig:polarization_measurement}(b), when measuring both linear and circular polarization components, QWPs are added to both the source and camera sides in addition to LPs. When $\theta_{\mathrm{L},p}$ and $\theta_{\mathrm{Q},p}$ represent the rotation angles of the LP and QWP on the source side, and $\theta_{\mathrm{L},a}$ and $\theta_{\mathrm{Q},a}$ represent those on the camera side, $\mathbf{P}$ and $\mathbf{A}$ can be expressed as:
\begin{align}
    \mathbf{P} &= \mathbf{Q}(\theta_{\mathrm{Q},p}) \mathbf{L}(\theta_{\mathrm{L},p}), \\
    \mathbf{A} &= \mathbf{L}(\theta_{\mathrm{L},a})\mathbf{Q}(\theta_{\mathrm{Q},a}),
\end{align}
where $\mathbf{Q}$ is the Mueller matrix of a QWP.

\begin{figure*}[t]
    \centering
    \includegraphics[width=\linewidth]{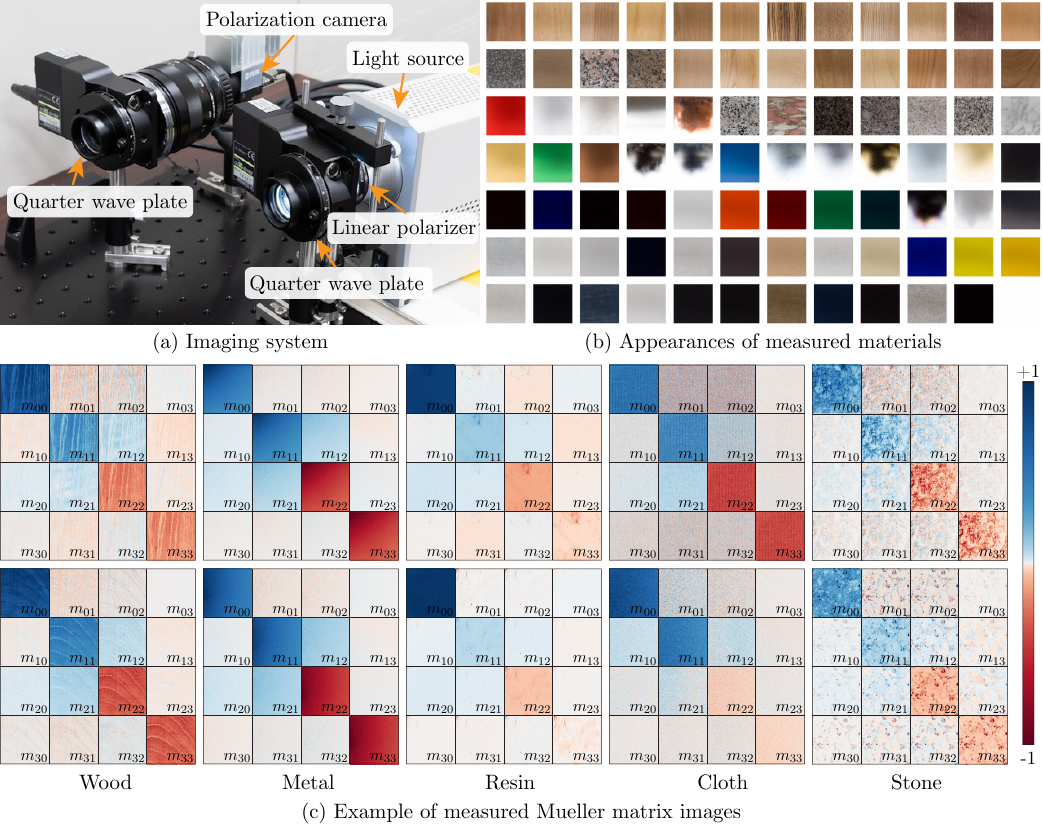}
    \caption{Our constructed Mueller-matrix material dataset. (a) The imaging system consists of a LP and a QWP in front of an unpolarized white light source, and a QWP in front of a linear polarization camera. Rotation stages rotate each QWP. (b) Appearance of the measured materials. (c) Example Mueller matrices for the measured materials. }
    \label{fig:detaset}
\end{figure*}

\section{Material Classification Based on Simultaneous Optimization}
\label{sec:simul-optim}

\begin{figure*}[t]
    \centering
    \includegraphics[width=\hsize]{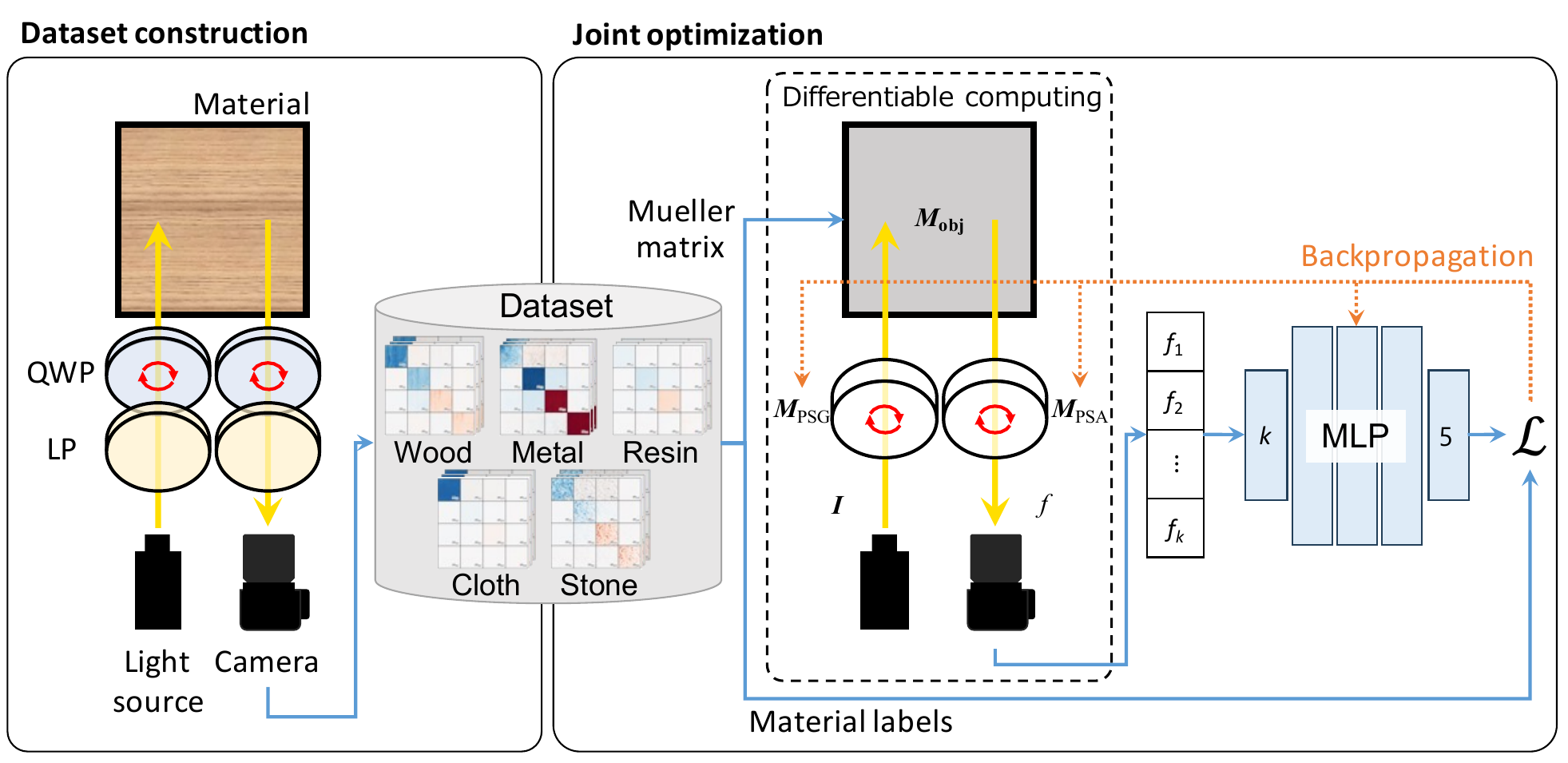}
    \caption{Overview of the ene-to-end optimization of polarization optics and material classifier. First, we measure Mueller matrices of various materials to construct a polarimetric reflectance dataset. Using this dataset, we perform joint optimization of the polarization element rotation angles and the material classifier (MLP).}
    \label{fig:our_methods}
\end{figure*}

Figure~\ref{fig:our_methods} shows an overview of our proposed simultaneous optimization approach. Since we aim to reduce the number of rotation angle combinations for LPs and QWPs, the Mueller matrix of target objects cannot be reconstructed from the reduced number of measurements. Therefore, instead of estimating the Mueller matrix for material classification, we adopt an end-to-end configuration where the set of light intensities $[f_1, f_2, \cdots, f_K]$ obtained by changing the rotation angles of polarization elements serves as input to the classifier. The light intensities for arbitrary rotation angles of polarization elements can be calculated using Equation~\eqref{eq:ellipsometer}. We calculate light intensities for various polarization element rotation angles and train the classifier to improve material recognition accuracy. During this process, we simultaneously optimize not only the classifier but also treat the combination of polarization element rotation angles as learnable parameters.

For the classifier, we use a multilayer perceptron (MLP) with ReLU activation functions, cross-entropy loss for multi-class classification, and the Adam optimizer~\cite{diederik2014adam}. The detaile of  the training are described in the appendix.

During training, we employed two types of data augmentation:
\begin{itemize}
    \item Random variations in light intensity
    \item Random rotations around the optical axis
\end{itemize}
For light intensity, we performed data augmentation by multiplying Mueller matrices with uniform random numbers ranging from $0.01$ to $10.0$. The rotation augmentation was implemented to improve model generalization, as polarization properties represented by Mueller matrices are affected by object orientation during measurement. The transformation of Mueller matrix components based on rotation can be performed using a rotation matrix $\mathbf{C}$.

\section{Experiments}

\subsection{Mueller-matrix Material Dataset}
Figure~\ref{fig:detaset} shows our constructed polarimetric reflectance properties dataset. Figure~\ref{fig:detaset}(a) shows the measurement apparatus, which is an ellipsometer consisting of LPs and QWPs as shown in Figure~\ref{fig:polarization_measurement}(b). The light source and camera are positioned as close as possible to create an approximate coaxial optical system.

The measurement targets are planar objects positioned to face the camera directly. As shown in Figure~\ref{fig:detaset}(b), we measured 83 different materials classified into five categories: wood, metal, resin, fabric, and stone. Figure~\ref{fig:detaset}(c) shows examples from the dataset, where each data point is a Mueller matrix image with Mueller matrix values stored for each pixel.

\subsection{Experimental Conditions}

\begin{figure*}[t]
    \centering
    \begin{minipage}{0.333\linewidth}
        \centering
        \includegraphics[width=\linewidth]{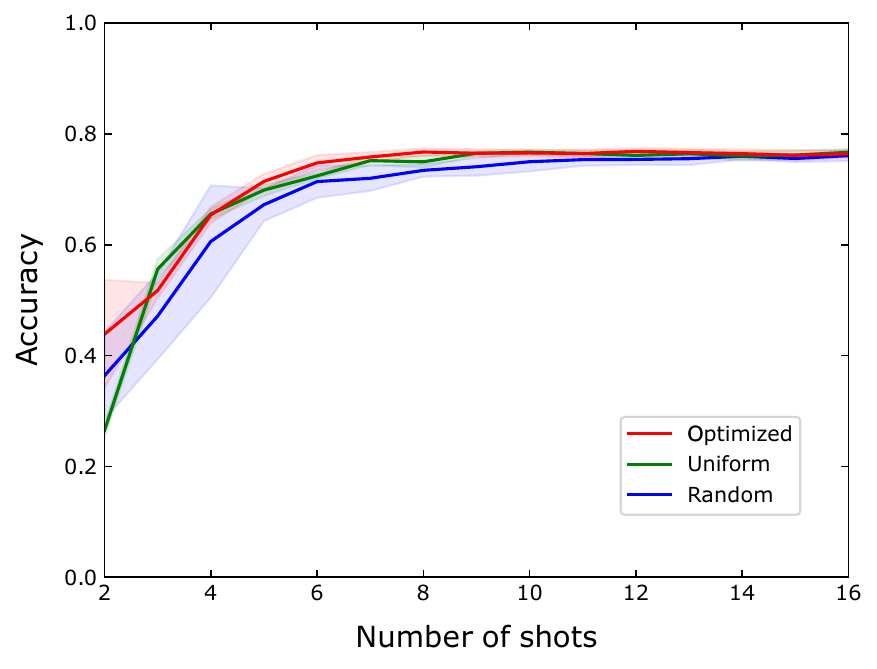}
        \textbf{LP:}\\Linear polarizer only
    \end{minipage}%
    \begin{minipage}{0.333\linewidth}
        \centering
        \includegraphics[width=\linewidth]{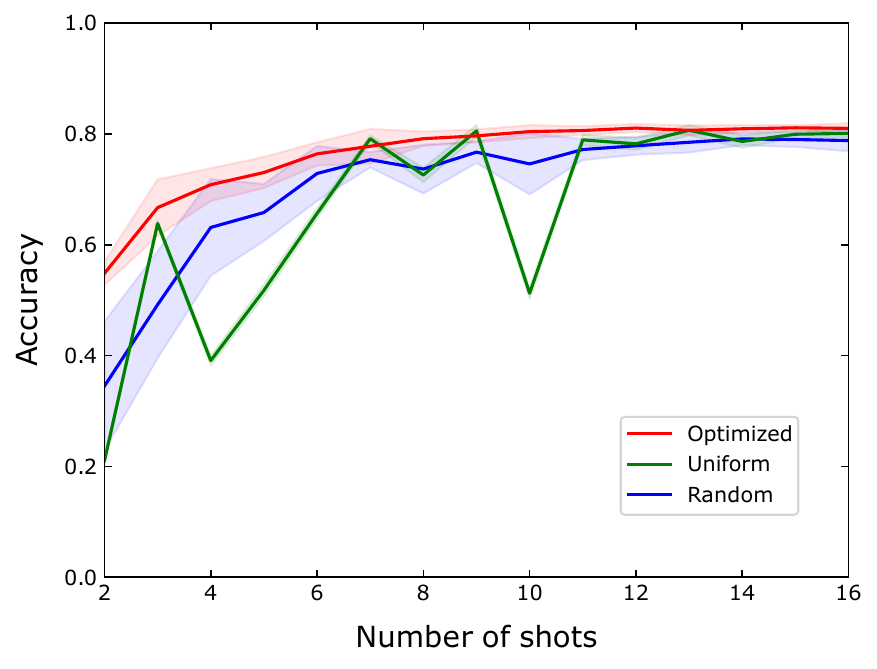}
        \textbf{QWP:}Rotates\\quarter-wave plates only
    \end{minipage}%
    \begin{minipage}{0.333\linewidth}
        \centering
        \includegraphics[width=\linewidth]{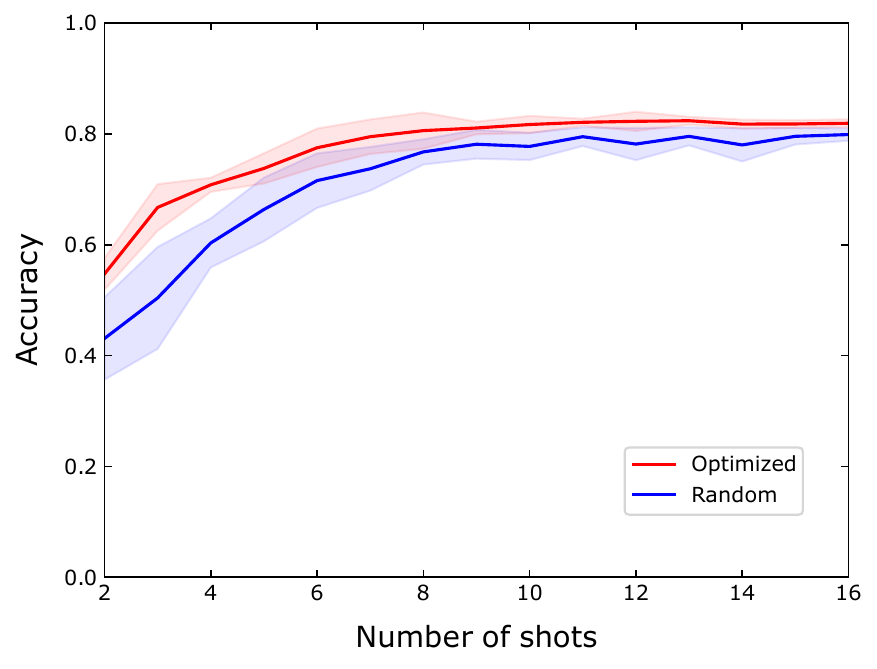}
        \textbf{LP+QWP:}Rotates linear polarizers and quarter-wave plates
    \end{minipage}
    \caption{Variation in classification accuracy with the number of captures for each polarization optical setup.}
    \label{fig:results-simOptim}
\end{figure*}

\begin{figure}[tb]
    \centering
    \includegraphics[width=0.9\hsize]{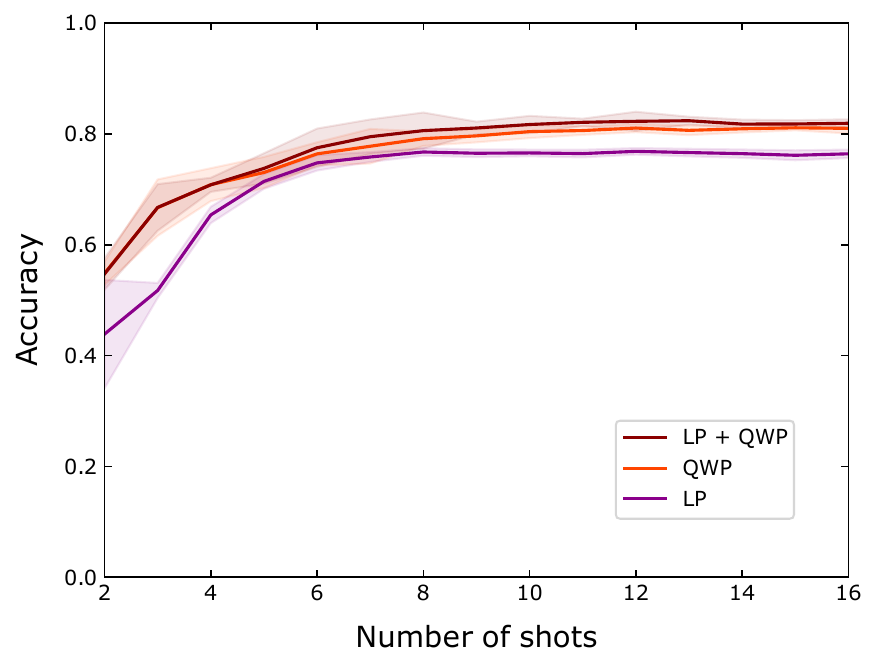}
    \caption{Comparison of classification accuracy of the proposed method across different polarization measurement setups.}
    \label{fig:acc_ours}
\end{figure}

We conducted material classification experiments under three different measurement conditions with varying degrees of control freedom for the polarization generator and analyzer configurations shown in Figure~\ref{fig:polarization_measurement}(a) and (b):
\vspace{1mm}\\
{
  \footnotesize
  \begin{tabular}{@{}lp{57mm}@{}}
    \toprule
    Condition & Description \\
    \midrule
    \textbf{LP} & Both polarization generator and analyzer consist of only linear polarizers that are rotated (Figure \ref{fig:polarization_measurement}(a)) \\
    \textbf{QWP} & Both polarization generator and analyzer consist of linear polarizers and quarter-wave plates, with only the quarter-wave plates being rotated (Figure \ref{fig:polarization_measurement}(b)) \\
    \textbf{LP+QWP} & Both polarization generator and analyzer consist of linear polarizers and quarter-wave plates, with both components being rotated (Figure \ref{fig:polarization_measurement}(b)) \\
    \bottomrule
  \end{tabular}
  \vspace{1mm}
}\\
To evaluate the effectiveness of simultaneous optimization of the polarization measurement system and classifier, we conducted training under the following conditions:
\vspace{1mm}\\
{
  \footnotesize
  \begin{tabular}{@{}lp{54mm}@{}}
    \toprule
    Optimization & Description \\
    \midrule
    \textbf{Random} & Only classifier optimization with random rotation of polarization elements \\
    \textbf{Uniform} & Only classifier optimization with systematic rotation of polarization elements \\
    \textbf{Optimized} & Joint optimization of both polarization elements and classifier through learning \\
    \bottomrule
  \end{tabular}
  \vspace{1mm}
}\\
Details of the \textbf{Uniform} optimization condition are provided in the appendix.

\subsection{Results}

Figure~\ref{fig:results-simOptim} shows the classification accuracy as a function of the number of measurements. 
To evaluate the robustness with respect to initialization, we conducted ten trials for both the \textbf{Random} and \textbf{Optimized} conditions. The solid lines represent the mean accuracy, while the shaded regions denote the standard deviation across trials. In all cases, our proposed simultaneous optimization method achieves higher accuracy with fewer measurements compared to random rotation configurations.
Additionally, under the \textbf{QWP-Uniform} condition, we observe significant drops in accuracy at certain measurement counts. This behavior can be attributed to redundancy in the measured polarization states, as discussed in Azzam’s method~\cite{baek2020image}. 

We compared the classification accuracy under different polarization measurement setups as shown in Figure~\ref{fig:acc_ours}. Two key observations can be drawn from these results. First, when using the QWP, the classification accuracy is higher than under the LP condition, indicating that circular polarization provides additional discriminative information for material recognition. Second, comparing the QWP and LP+QWP configurations, we observe a slight improvement in accuracy for LP+QWP, suggesting that a higher degree of freedom in polarization modulation leads to more effective optical configurations for material classification.

\section{Discussion}

The experimental results presented in the previous section demonstrate that classification accuracy improves when the polarization optical system is optimized jointly with the classifier. Beyond performance gains, analyzing the optimized system parameters can provide valuable insights into the relationship between polarization characteristics and material classification. In this section, we discuss the combinations of rotation angles obtained through optimization for the LPs and QWPs.

\subsection{Case with Rotating LPs Only}

As described in Section~\ref{sec:simul-optim}, during training, data augmentation involves randomly rotating the target around the optical axis. Consequently, the optimized angles of the linear polarizers are not fixed and vary randomly across samples. Moreover, the sequence in which combinations of polarizer angles appear over multiple measurements is also randomized. 

\begin{figure*}[t]
    \centering
    \begin{minipage}{0.48\hsize}
        \centering
        \includegraphics[width=\hsize]{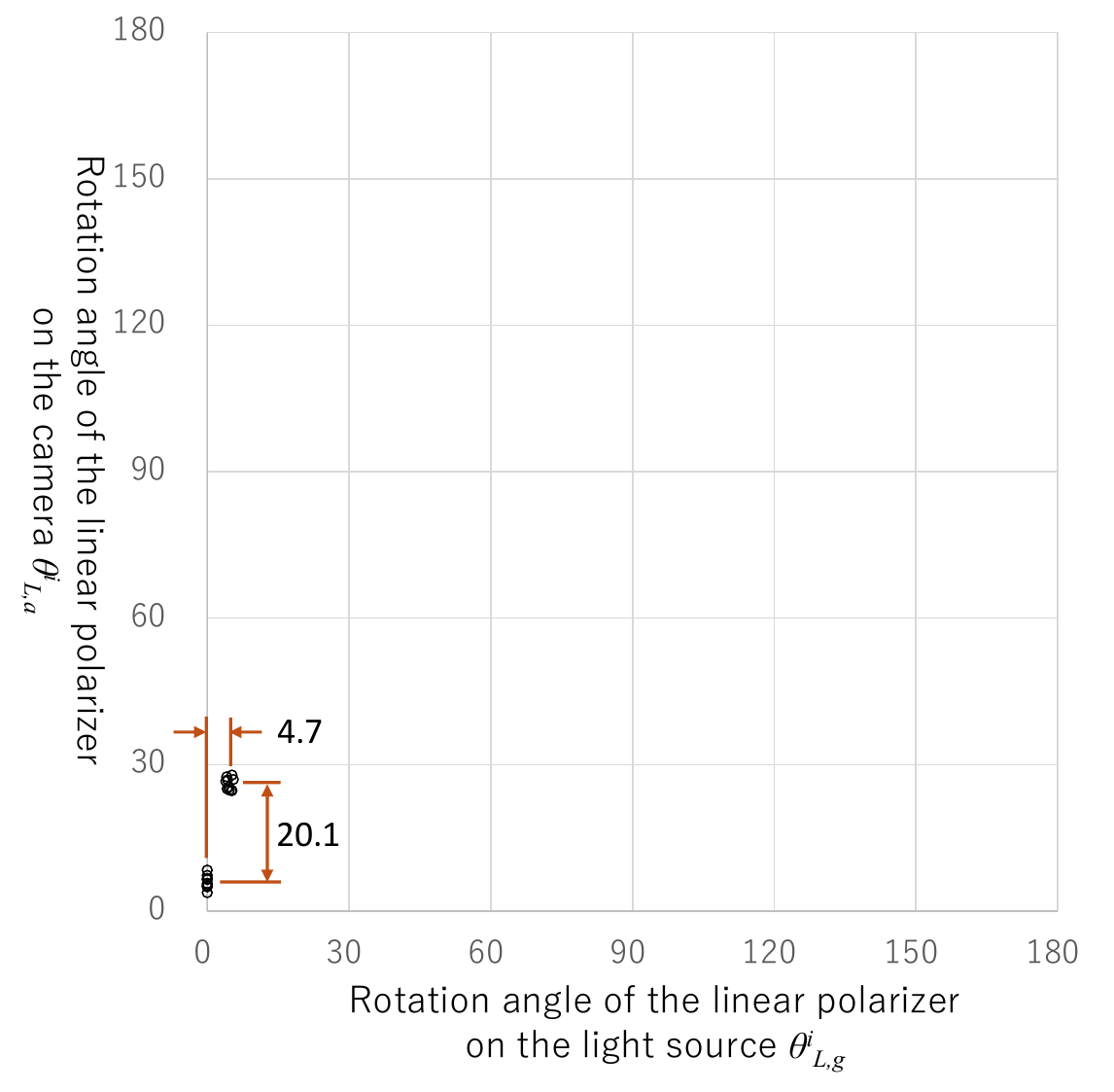}\\(a) $K=2$ measurements
    \end{minipage}
    \begin{minipage}{0.48\hsize}
        \centering
        \includegraphics[width=\hsize]{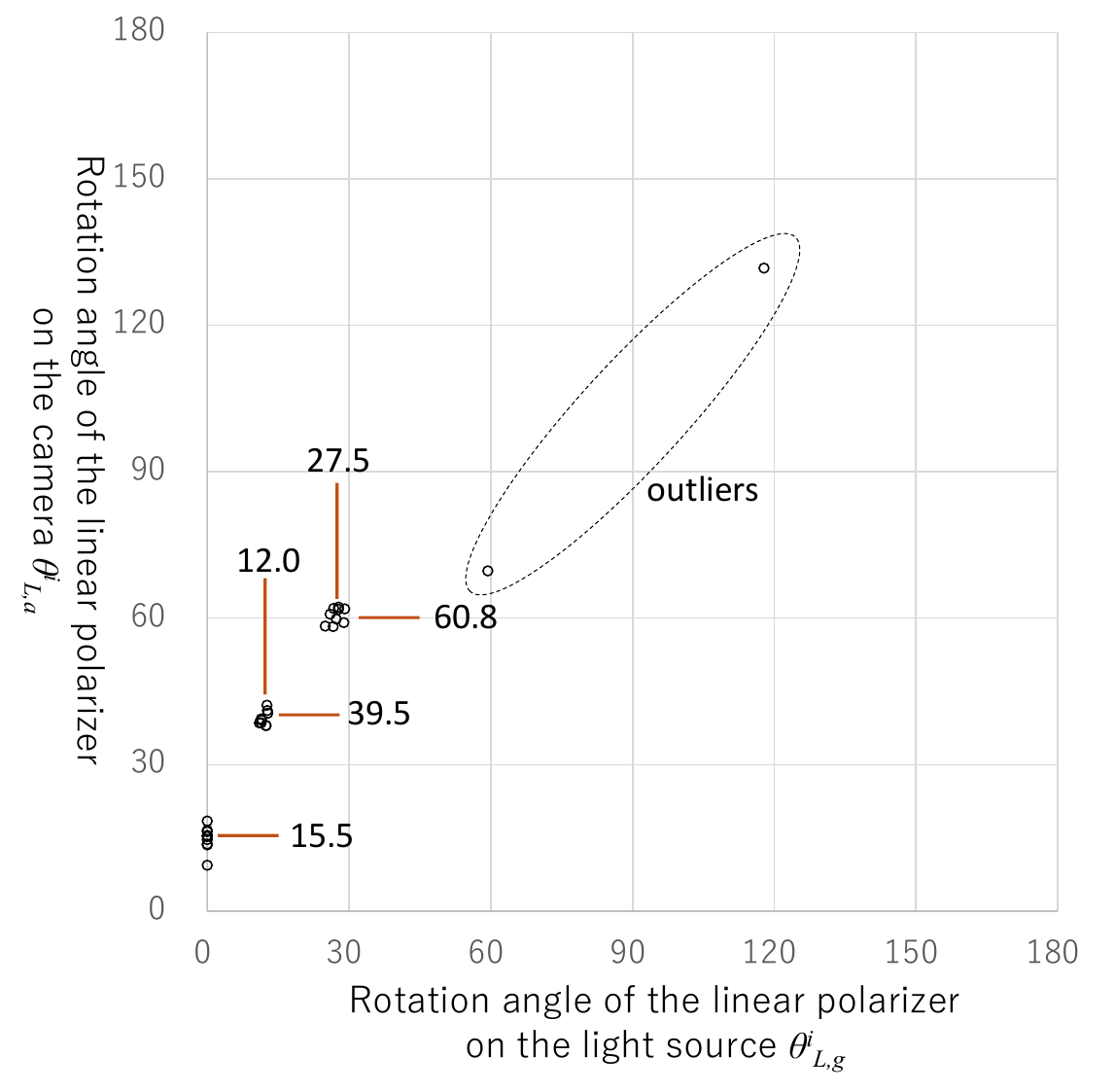}\\(b) $K=3$ measurements
    \end{minipage}
    \caption{Distribution of rotation angles when using only a linear polarizer (LP condition).}
    \label{fig:angle-distri-linear}
\end{figure*}

To enable an interpretable analysis, we remove this randomness through the following procedure. First, we define $\theta_{\mathrm{L},g}^i$ and $\theta_{\mathrm{L},a}^i$ as the optimized rotation angles of the LPs on the light source and camera sides during the $i$-th measurement, respectively, and calculate their relative angles
$\theta_{\mathrm{L},d}^i = \theta_{\mathrm{L},g}^i - \theta_{\mathrm{L},a}^i$.
Next, we use the angle of the light source polarizer at measurement where the relative angle is minimum $i_{\mathrm{min}} = \underset{i} {\operatorname{argmin}}(\theta_{\mathrm{L},d}^i)$ as a reference, and determine the distribution of relative rotation angles for each linear polarizer. The resulting scatter plot is shown in Figure~\ref{fig:angle-distri-linear}.

\begin{figure*}[t]
    \centering
    \begin{minipage}{0.48\hsize}
        \centering
        \includegraphics[width=\hsize]{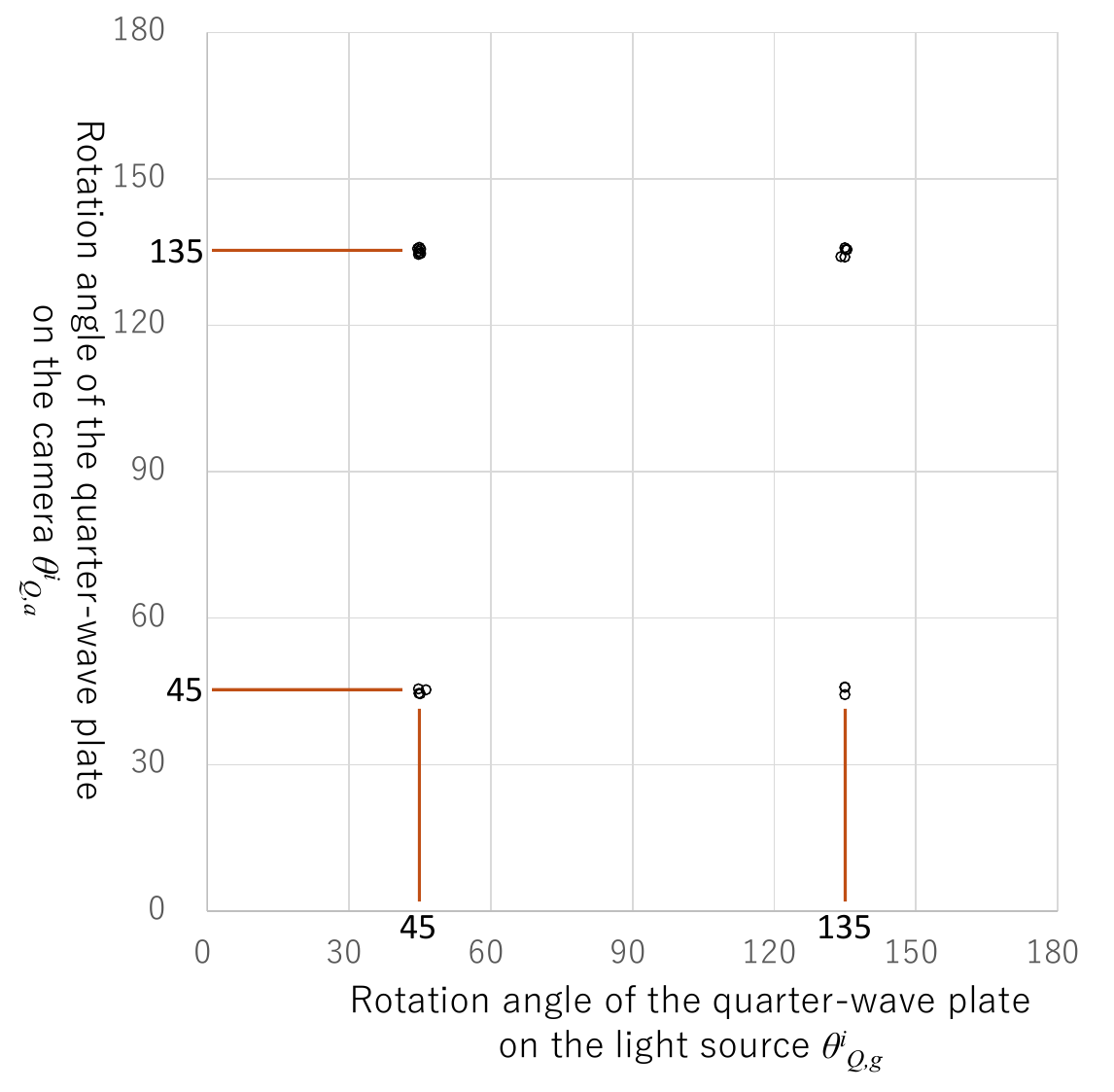}\\(a) $K=2$ measurements
    \end{minipage}
    \begin{minipage}{0.48\hsize}
        \centering
        \includegraphics[width=\hsize]{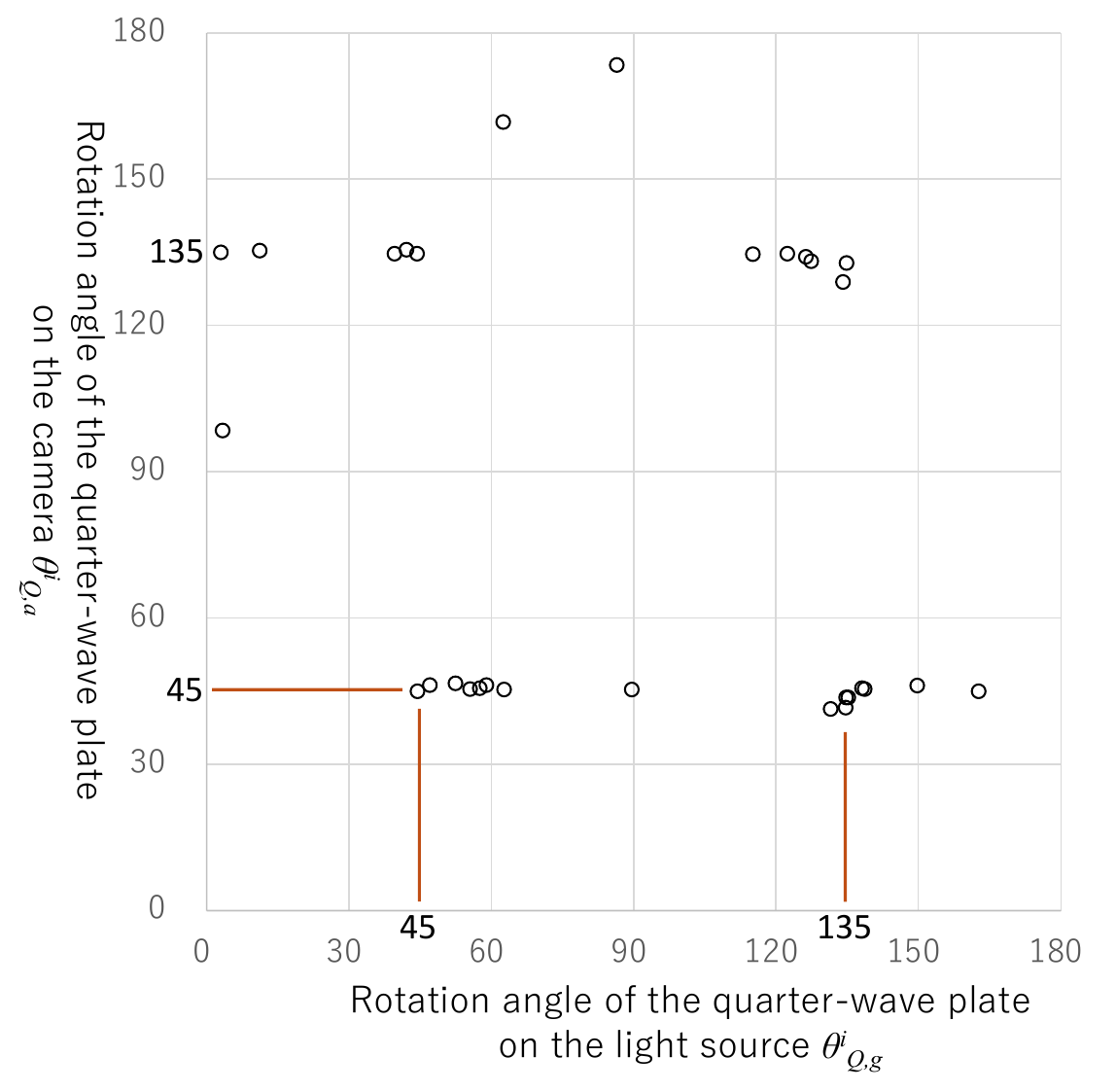}\\(b) $K=3$ measurements
    \end{minipage}
    \caption{Distribution of rotation angles when only the quarter-wave plate is rotated (QWP condition).}
    \label{fig:angle-distri-retarder}
\end{figure*}

In the two-measurement case ($K = 2$), as shown in Figure~\ref{fig:angle-distri-linear}(a), we observe two distinct measurement configurations in which the relative angles between the light-source and camera-side polarizers are either small (average of $6.0^{\circ}$) or large (average of $26.1^{\circ}$). Notably, the difference between these two average angles ($26.1^{\circ} - 6.0^{\circ} = 20.1^{\circ}$) does not correspond to any conventional configurations—such as parallel ($0^{\circ}$) or crossed-Nicol ($90^{\circ}$)—indicating that the optimized results diverge from intuitive or empirically designed polarizer arrangements.

Furthermore, the rotation angle of the light-source polarizer is slightly different (approximately $4.7^{\circ}$). Such behavior would be difficult to derive through empirical design alone. The rotation angles obtained from all trials form compact clusters with low variance, suggesting that the optimization process converged reliably.

In the three-measurement ($K=3$) case shown in Figure~\ref{fig:angle-distri-linear}(b), the relative angles between the light source and camera side polarizers form three clusters with averages of $15.5^{\circ}, 39.5^{\circ},$ and $60.8^{\circ}$ in ascending order, and the corresponding rotation angles of the light source polarizer show non-uniform shifts of $12.0^{\circ}$ and $27.5^{\circ}$ relative to the reference. Among the ten trials, only one significantly deviated from these distributions, likely due to insufficient optimization or convergence to a local minimum, but the optimization appears to have converged sufficiently in all other cases.

\subsection{Case with Rotating QWPs}
When only the QWPs are rotated, the polarization planes of the LPs on both the light-source and camera sides are fixed vertically, serving as a reference. This configuration removes the angular randomness discussed in the previous section. Consequently, as shown in Figure~\ref{fig:angle-distri-retarder}, we directly present a scatter plot of the optimized angles of the QWPs.

In the two-measurement case ($K = 2$), shown in Figure~\ref{fig:angle-distri-retarder}(a), the optimization yields configurations where the retardation axes of the QWPs are oriented at either $45^{\circ}$ or $135^{\circ}$ relative to the polarizers on both the light-source and camera sides. These configurations correspond to the generation of circularly polarized light on the illumination side and the detection of right- and left-handed circular polarization on the observation side. Unlike linear polarization, circular polarization remains invariant to rotations of the target object around the optical axis, suggesting that it serves as an effective and stable feature for material classification.

Similarly, in the three-measurement case ($K = 3$), shown in Figure~\ref{fig:angle-distri-retarder}(b), the optimized angles of the camera-side QWP are again concentrated around $45^{\circ}$ and $135^{\circ}$, reinforcing the effectiveness of circular polarization detection. In contrast, the light-source-side QWP exhibits slightly greater variation, though its optimal angles are still predominantly distributed near $45^{\circ}$ and $135^{\circ}$.

% These findings, along with the results shown in Figure~\ref{fig:acc_ours}, suggest that the use of circular polarization (both in illumination and detection) is effective for material classification.

\section{Conclusion}

In this study, we proposed a method for material classification based on the end-to-end joint optimization of polarizing element rotation angles in a polarization measurement system and a material classifier. We conducted training under two conditions: (1) using only LP and (2) using both LP and QWP. In both settings, the proposed method achieved high classification accuracy with a small number of measurements. Moreover, the jointly optimized rotation angles of both the LPs and QWPs achieved superior classification performance compared to the cases where only the LPs or only the QWPs.

% \todo{Limitation: No color/texture information used. Limited to RGB (wavelength dependency). Only frontal material surfaces.}

\textbf{Limitations:} 
Despite its promising results, our method has several limitations. 
First, the experiments were conducted on planar materials under controlled laboratory conditions, and their robustness to complex real-world scenes with spatially varying geometry and illumination remains to be validated. 
Second, our approach relies solely on polarization cues without incorporating complementary information such as color, texture, or spectral reflectance, which may further enhance recognition performance. 

\section*{\uppercase{Appendix}}
\appendix
\section{Mueller Matrices}
Here we present representative Mueller matrices.

The Mueller matrix $\mathbf{L}\,(\theta)$ of a linear polarizer with rotation angle $\theta$ is expressed as follows:
\begin{eqnarray}
    \resizebox{0.82\hsize}{!}{$%
        \mathbf{L}\,(\theta)\!=\!\dfrac{1}{2}\!
        \begin{bmatrix}
        1 & \cos2\theta & \sin2\theta & 0 \\
        \cos2\theta & \cos^22\theta & \sin2\theta\cos2\theta & 0 \\
        \sin2\theta & \sin2\theta\cos2\theta & \sin^22\theta & 0 \\
        0 & 0 & 0 & 0
        \end{bmatrix}%
    $}.
    \label{eqpre:1}
\end{eqnarray}

The Mueller matrix $\mathbf{Q}\,(\theta)$ of a quarter-wave plate with rotation angle $\theta$ is expressed as follows:
\begin{eqnarray}
    \resizebox{0.82\hsize}{!}{$%
        \mathbf{Q}\,(\theta)\!=\!
        \begin{bmatrix}
        1 & 0 & 0 & 0 \\
        0 & \cos^22\theta & \sin2\theta\cos2\theta & -\sin2\theta \\
        0 & \sin2\theta\cos2\theta & \sin^22\theta & \cos2\theta \\
        0 & \sin2\theta & -\cos2\theta & 0
        \end{bmatrix}%
    $}.
    \label{eqpre:2}
\end{eqnarray}

The rotation matrix $\mathbf{C}\,(\theta)$ used when rotating by angle $\theta$ around the optical axis from the original angle is expressed as follows:
\begin{eqnarray}
    \mathbf{C}\,(\theta)=
    \begin{bmatrix}
    1 & 0 & 0 & 0 \\
    0 & \cos2\theta & \sin2\theta & 0 \\
    0 & -\sin2\theta & \cos2\theta & 0 \\
    0 & 0 & 0 & 1
    \end{bmatrix}.
    \label{eqpre:3}
\end{eqnarray}
Using this rotation matrix, the Mueller matrix $\mathbf{M}(\theta)$ rotated by angle $\theta$ can be calculated as follows:
\begin{eqnarray}
    \mathbf{M}(\theta)=\mathbf{C}\,(-\theta)\mathbf{M}\mathbf{C}\,(\theta). 
    \label{eqpre:4}
\end{eqnarray}

\section{Training Details}
% \todo{Describe training details, network method, optimizer, bias handling, etc.}

Table~\ref{tab:mlp-architecture} shows the architeture of material classifier. 
During training, we uniformly sampled 128 Mueller matrices as mini-batches from the dataset of Mueller matrix images.
Since there is an imbalance in the number of material categories, we corrected this bias by calculating weights based on the number of categories in each mini-batch when computing the loss. 
% \todo{Categories?}

\begin{table}[t]
  \centering
  \caption{Architecture of the  MLP used in our method. Each linear layer is followed by a ReLU activation function, except for the output layer.}
  \label{tab:mlp-architecture}
  \begin{tabular}{@{}ccc@{}}
    \toprule
    Layer & Input Size & Output Size \\
    \midrule
    Linear + ReLU & $C_{\text{in}}$ & 64 \\
    Linear + ReLU & 64 & 32 \\
    Linear + ReLU & 32 & 32 \\
    Linear + ReLU & 32 & 16 \\
    Linear & 16 & $C_{\text{out}}$ \\
    \bottomrule
  \end{tabular}
\end{table}

The rotation angles of polarizing elements under the \textbf{Uniform} optimization condition were determined as shown in Table \ref{tab:uniform-condition}.

\begin{table}[t]
  \centering
  \caption{Details of uniform measurements configuration.}
  \label{tab:uniform-condition}

  \footnotesize
  \begin{tabular}{@{}lp{41mm}@{}}
    \toprule
    Condition Combination & Description \\
    \midrule
    \textbf{LP-Uniform} & Following the four directional polarizers mounted in polarization cameras, we determined combinations of azimuth angles $\theta_{\mathrm{L},g}$ and $\theta_{\mathrm{L},a}$ from $0^{\circ}$, $45^{\circ}$, $90^{\circ}$, and $135^{\circ}$. \\
    \textbf{QWP-Uniform} & Based on the double rotating compensator method proposed by Azzam et al. \cite{azzam1978photopolarimetric}, we determined azimuth angle combinations such that $\theta_{\mathrm{Q},a}\!=\!5\theta_{\mathrm{Q},g}$. \\
    \textbf{LP+QWP-Uniform} & We did not conduct experiments as there are no precedents for methods that perform optimal measurements while independently rotating four polarizing optical elements. \\
    \bottomrule
  \end{tabular}
  \vspace{2mm}
\end{table}

\bibliographystyle{apalike}
\bibliography{references}

\end{document}